\documentclass[letterpaper, 10 pt, conference]{ieeeconf}  

\usepackage{amsfonts}
\usepackage{amsmath}
\usepackage{amssymb}
\usepackage{dsfont}
\usepackage{eucal}
\usepackage{float}
\usepackage[edges]{forest}
\usepackage{listings}
\usepackage{mathtools}
\usepackage[short]{optidef}
\usepackage{subcaption}  
\usepackage{tikz}
\usepackage{url}
\usepackage{xcolor}

\newtheorem{problem}{Problem}
\newtheorem{remark}{Remark}

\definecolor{folderbg}{RGB}{124,166,198}
\definecolor{folderborder}{RGB}{110,144,169}
\definecolor{myLightGray}{RGB}{191,191,191}
\definecolor{myGray}{RGB}{160,160,160}
\definecolor{myDarkGray}{RGB}{144,144,144}
\definecolor{myDarkRed}{RGB}{167,114,115}
\definecolor{myLightRed}{RGB}{255,114,118}
\definecolor{myRed}{RGB}{255,58,70}
\definecolor{myGreen}{RGB}{0,200,71}
\definecolor{myBlue}{RGB}{100,149,237}
\newlength\Size
\tikzset{%
  folder/.pic={%
    \filldraw [draw=folderborder, top color=folderbg!50, bottom color=folderbg] (-1.05*\Size,0.05\Size+5pt) rectangle ++(.75*\Size,-0.05\Size-5pt);
    \filldraw [draw=folderborder, top color=folderbg!50, bottom color=folderbg] (-1.15*\Size,-0.85*\Size) rectangle (1.15*\Size,0.85*\Size);
  },
  file/.pic={%
    \filldraw [draw=folderborder, top color=folderbg!5, bottom color=folderbg!10] (-\Size,.4*\Size+5pt) coordinate (a) |- (\Size,-1.2*\Size) coordinate (b) -- ++(0,1.6*\Size) coordinate (c) -- ++(-5pt,5pt) coordinate (d) -- cycle (d) |- (c) ;
  },
}
\forestset{%
  declare autowrapped toks={pic me}{},
  pic dir tree/.style={%
    for tree={%
      folder,
      grow'=0,
      font=\small,
    },
    before typesetting nodes={%
      for tree={%
        edge label+/.option={pic me},
      },
    },
  },
  pic me set/.code n args=2{%
    \forestset{%
      #1/.style={%
        inner xsep=2\Size,
        inner ysep=0\Size,
        pic me={pic {#2}},
      }
    }
  },
  pic me set={directory}{folder},
  pic me set={file}{file},
}
\renewcommand{\vec}[1]{\textbf{#1}}

\IEEEoverridecommandlockouts

\title{\LARGE \bf
Decentralized Vehicle Coordination: The Berkeley DeepDrive Drone Dataset and Consensus-Based Models
}

\author{Fangyu Wu$^{1,2}$, Dequan Wang$^{2}$, Minjune Hwang$^{2}$, Chenhui Hao$^{3}$, Jiawei Lu$^{2}$, Jiamu Zhang$^{2}$, Christopher Chou$^{2}$, \\Trevor Darrell$^{2}$, and Alexandre Bayen$^{2}$
\thanks{*This work was supported by Berkeley DeepDrive.}
\thanks{$^{1}$Department of Electrical and Computer Engineering at Cornell University, $^{2}$Department of Electrical Engineering and Computer Sciences at the University of California, Berkeley, and $^{3}$Department of Civil and Environmental Engineering at the University of California, Berkeley}%
}

\begin{document}

\makeatletter
\g@addto@macro\@maketitle{
  \begin{figure}[H]
  \setlength{\linewidth}{\textwidth}
  \setlength{\hsize}{\textwidth}
  \centering
  \begin{subfigure}[t]{0.11\textwidth}
      \centering
      \includegraphics[width=\textwidth]{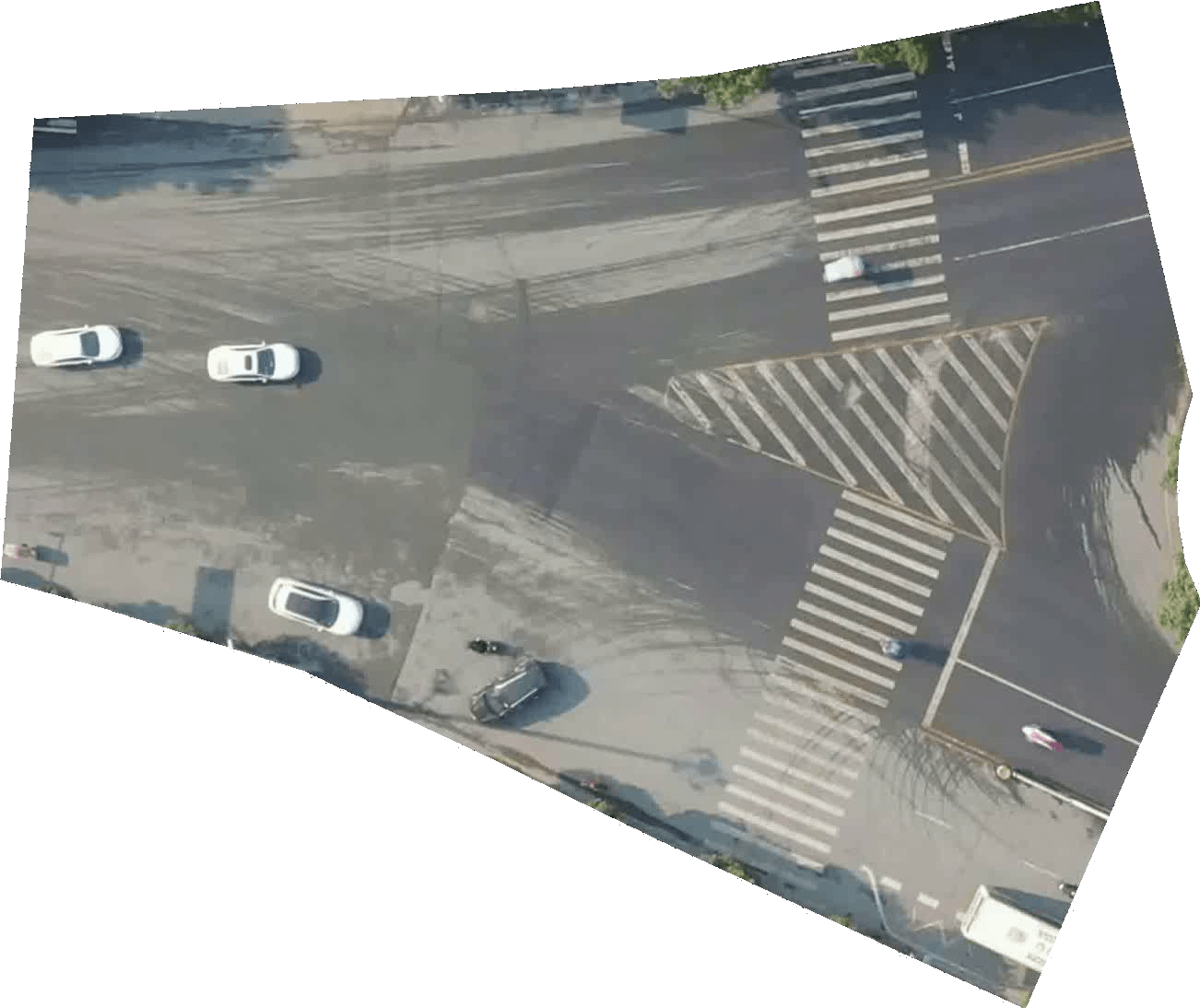}
      \caption{jnc00.mp4}
      \label{subfig:jnc00}
  \end{subfigure}
  \hfill
  \begin{subfigure}[t]{0.16\textwidth}
      \centering
      \includegraphics[width=\textwidth]{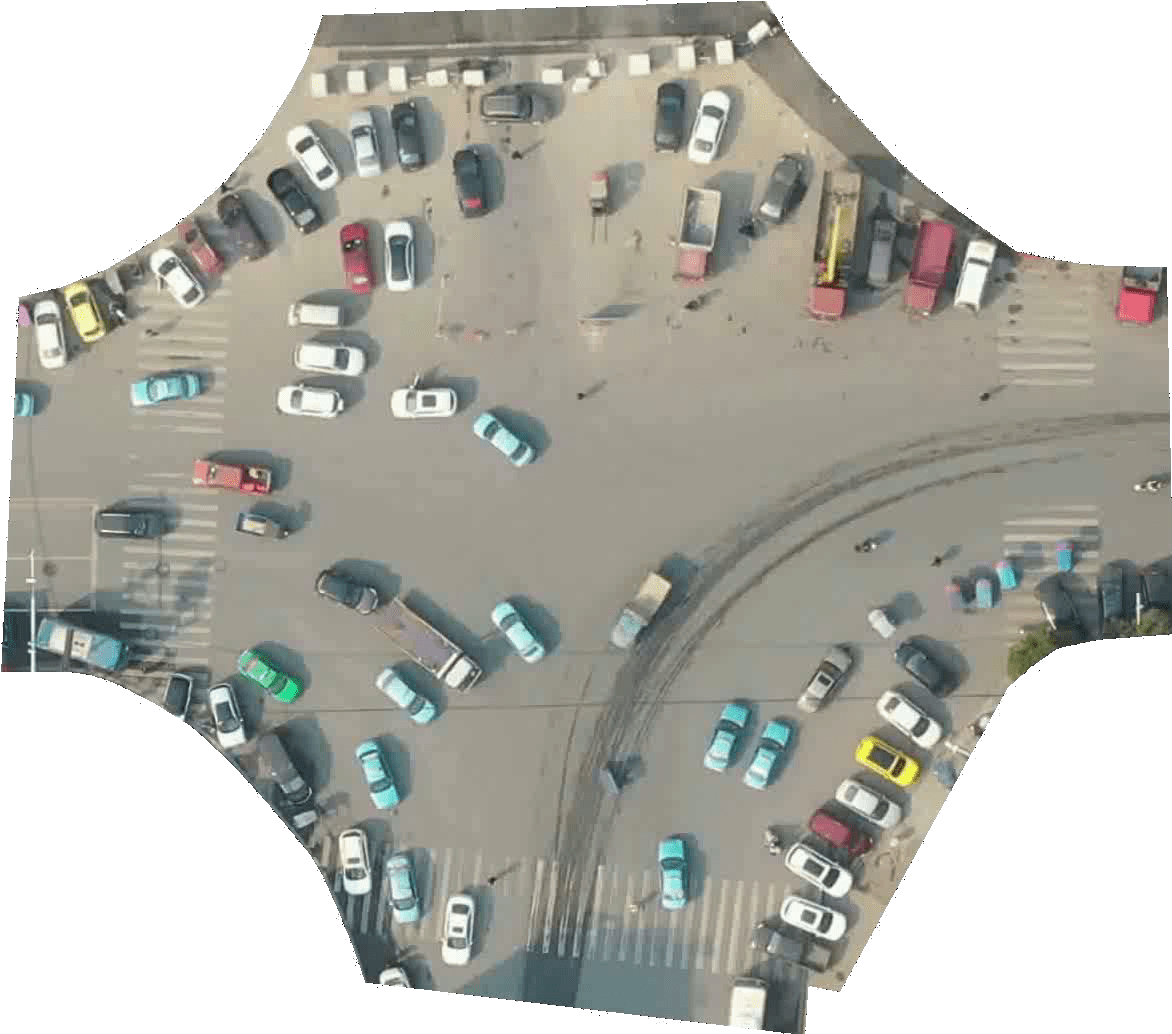}
      \caption{jnc01.mp4}
      \label{subfig:jnc01}
  \end{subfigure}
  \hfill
  \begin{subfigure}[t]{0.1\textwidth}
      \centering
      \includegraphics[width=\textwidth]{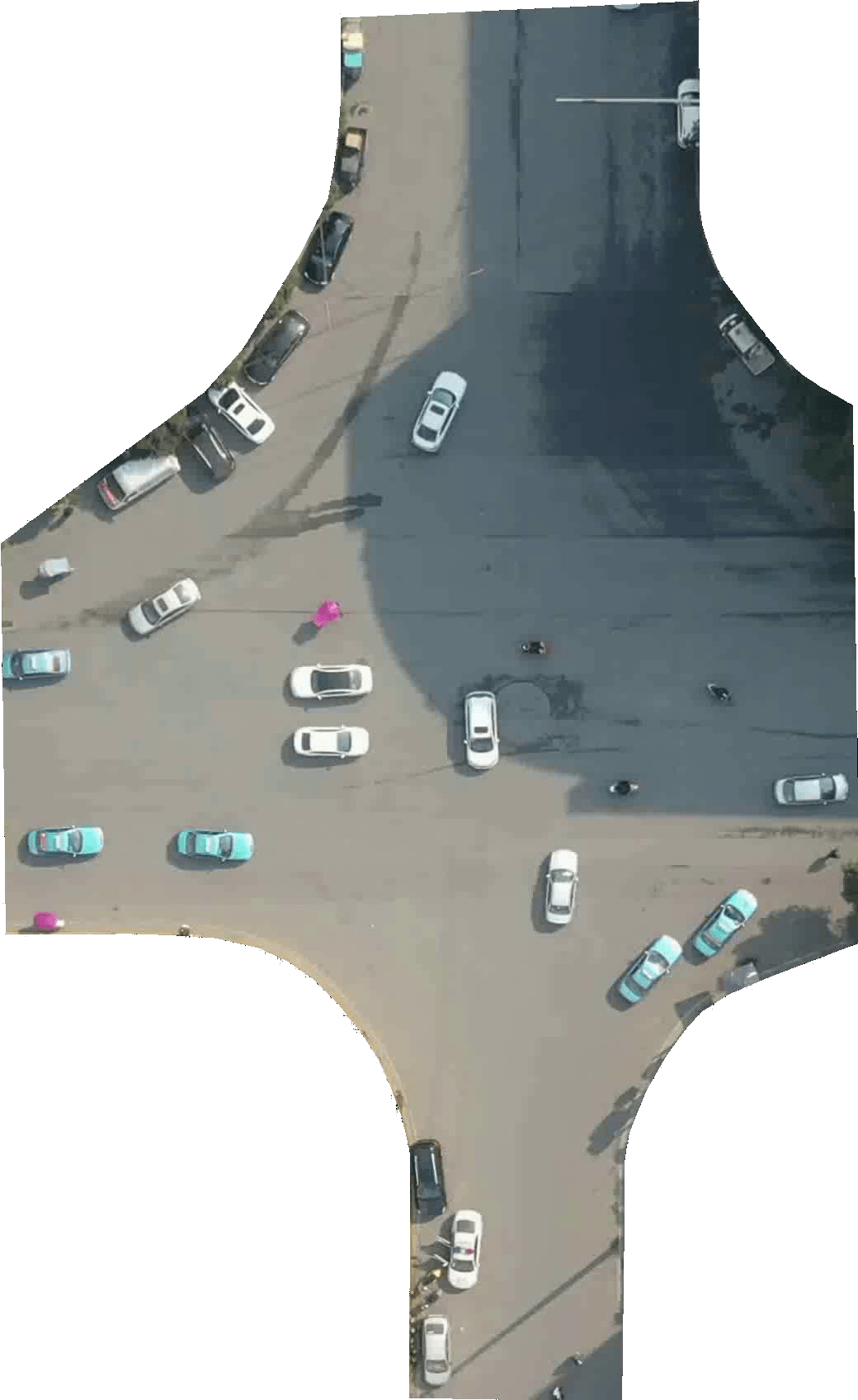}
      \caption{jnc02.mp4}
      \label{subfig:jnc02}
  \end{subfigure}
  \hfill
  \begin{subfigure}[t]{0.115\textwidth}
      \centering
      \includegraphics[width=\textwidth]{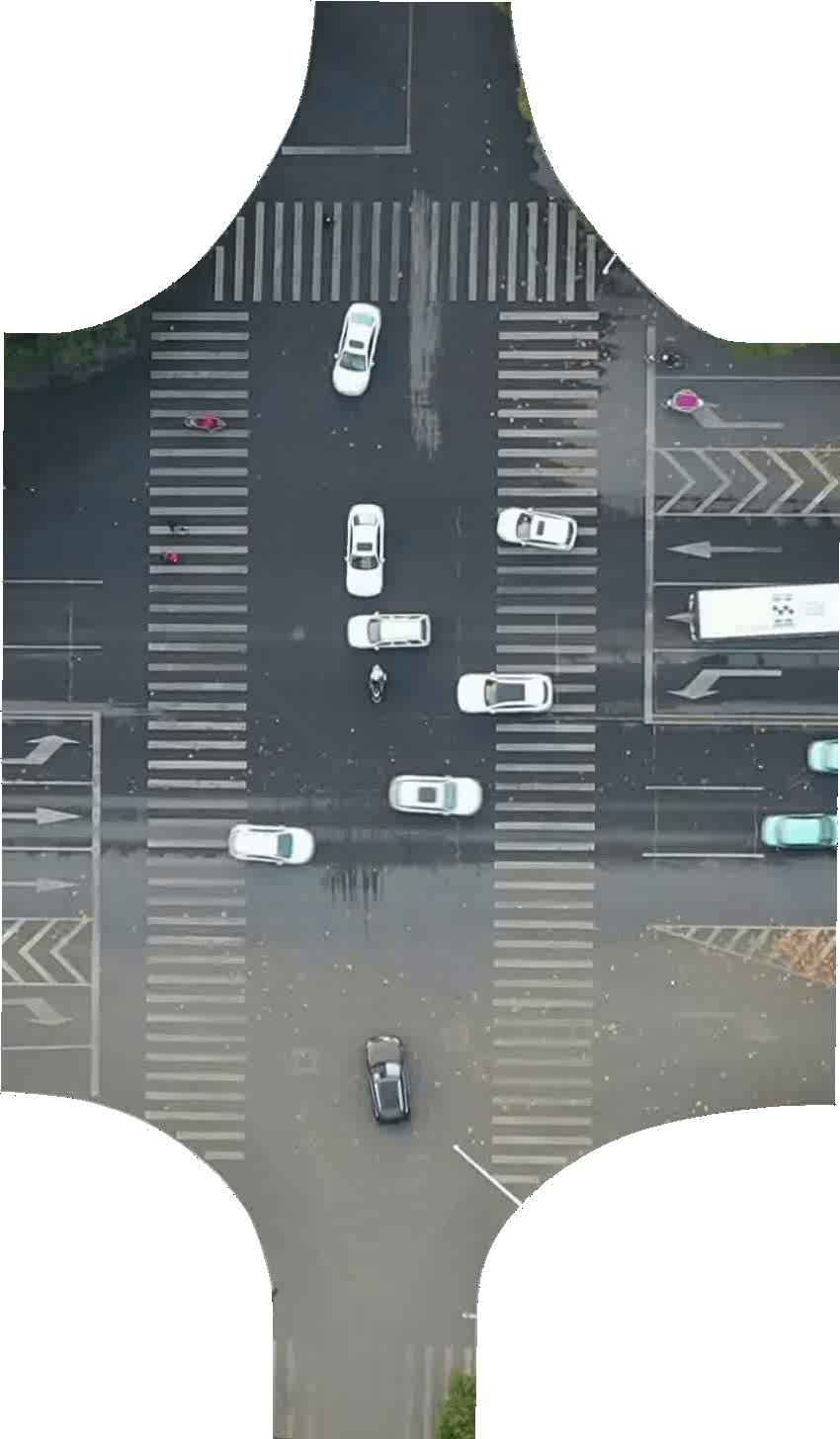}
      \caption{jnc07.mp4}
      \label{subfig:jnc07}
  \end{subfigure}
  \hfill
  \begin{subfigure}[t]{0.19\textwidth}
      \centering
      \includegraphics[width=\textwidth]{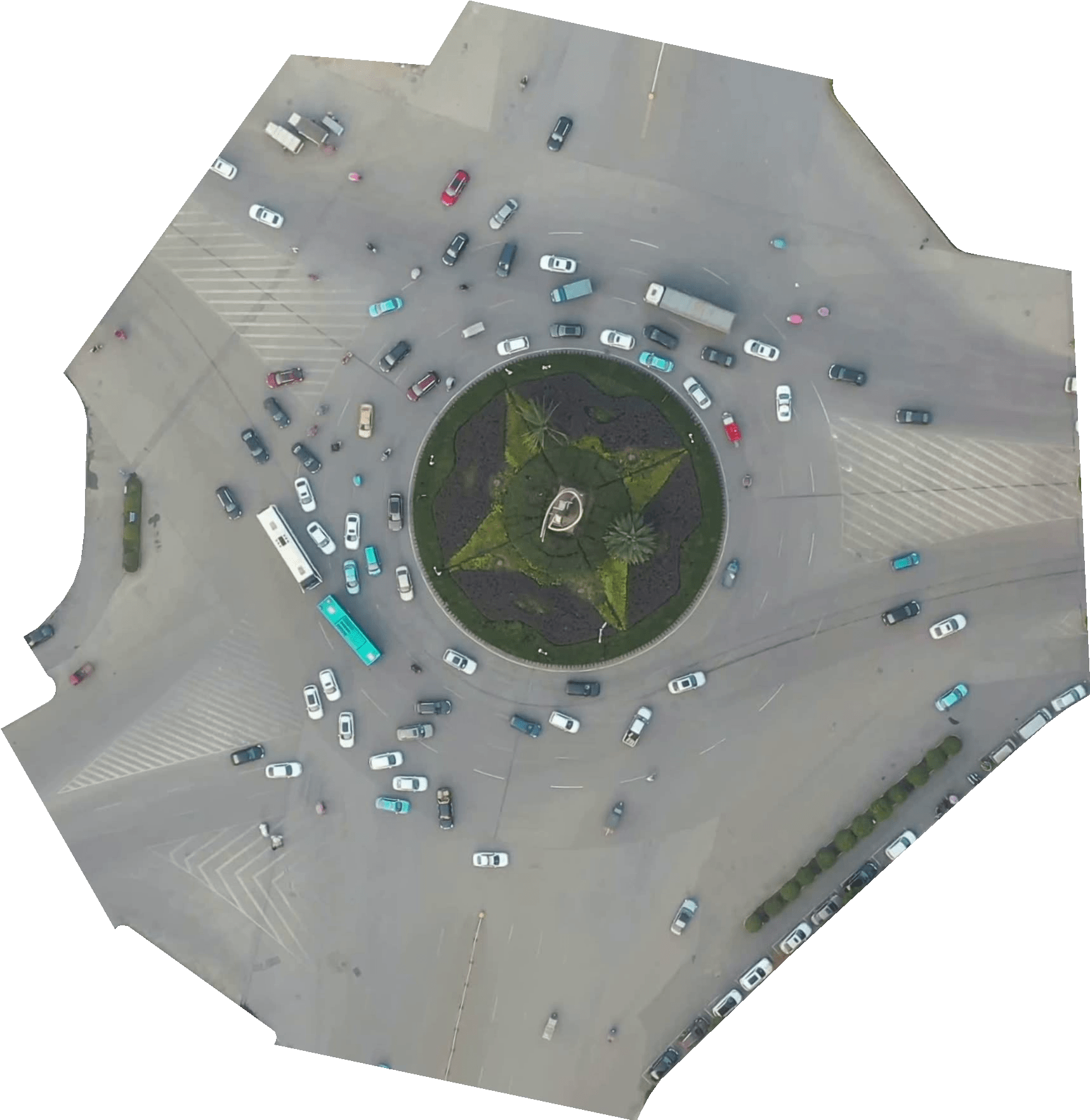}
      \caption{jnc03.mp4, jnc04.mp4}
      \label{subfig:jnc04}
  \end{subfigure}
  \hfill
  \begin{subfigure}[t]{0.14\textwidth}
      \centering
      \includegraphics[width=\textwidth]{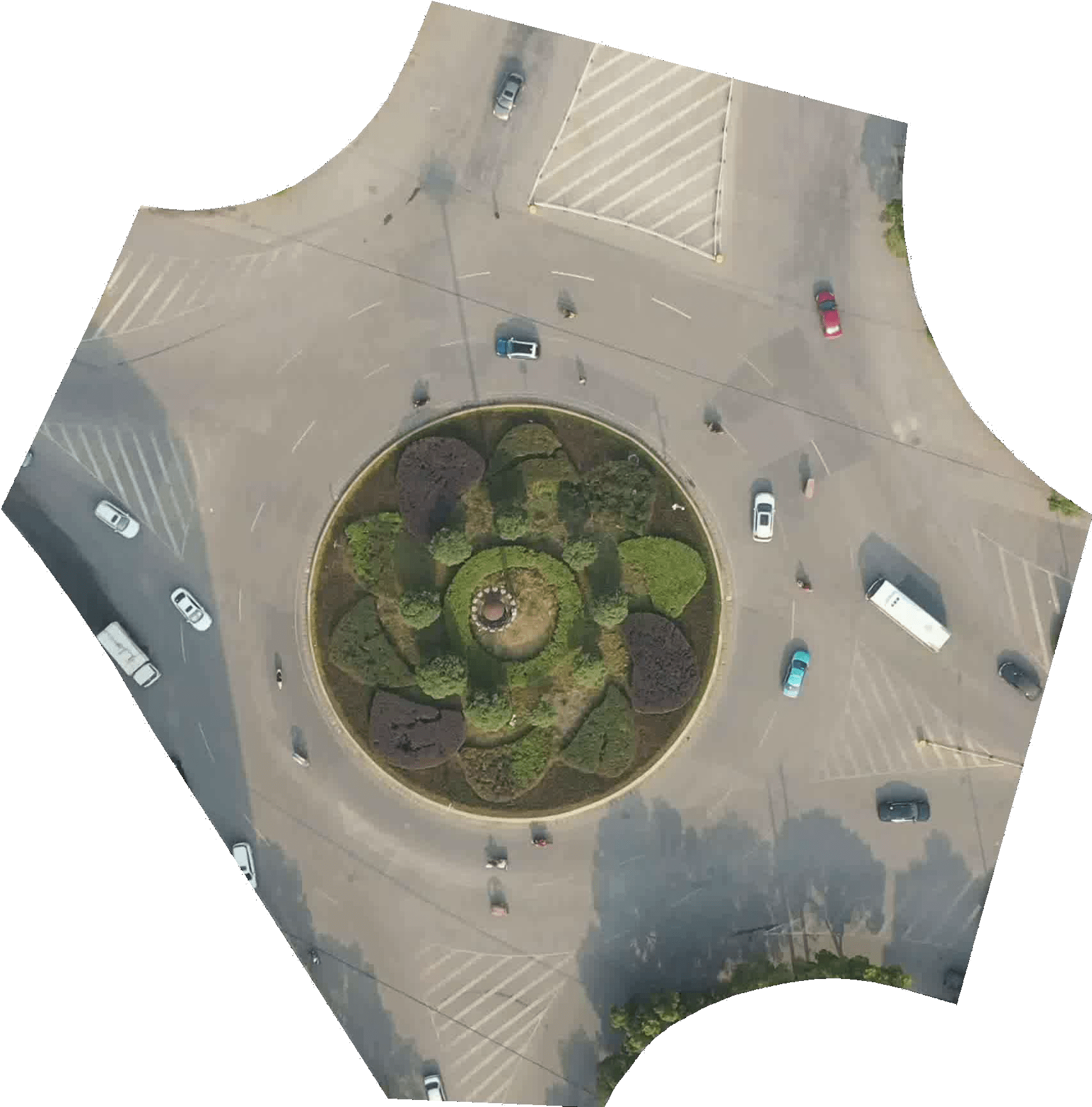}
      \caption{jnc05.mp4}
      \label{subfig:jnc05}
  \end{subfigure}
  \hfill
  \begin{subfigure}[t]{0.11\textwidth}
      \centering
      \includegraphics[width=\textwidth]{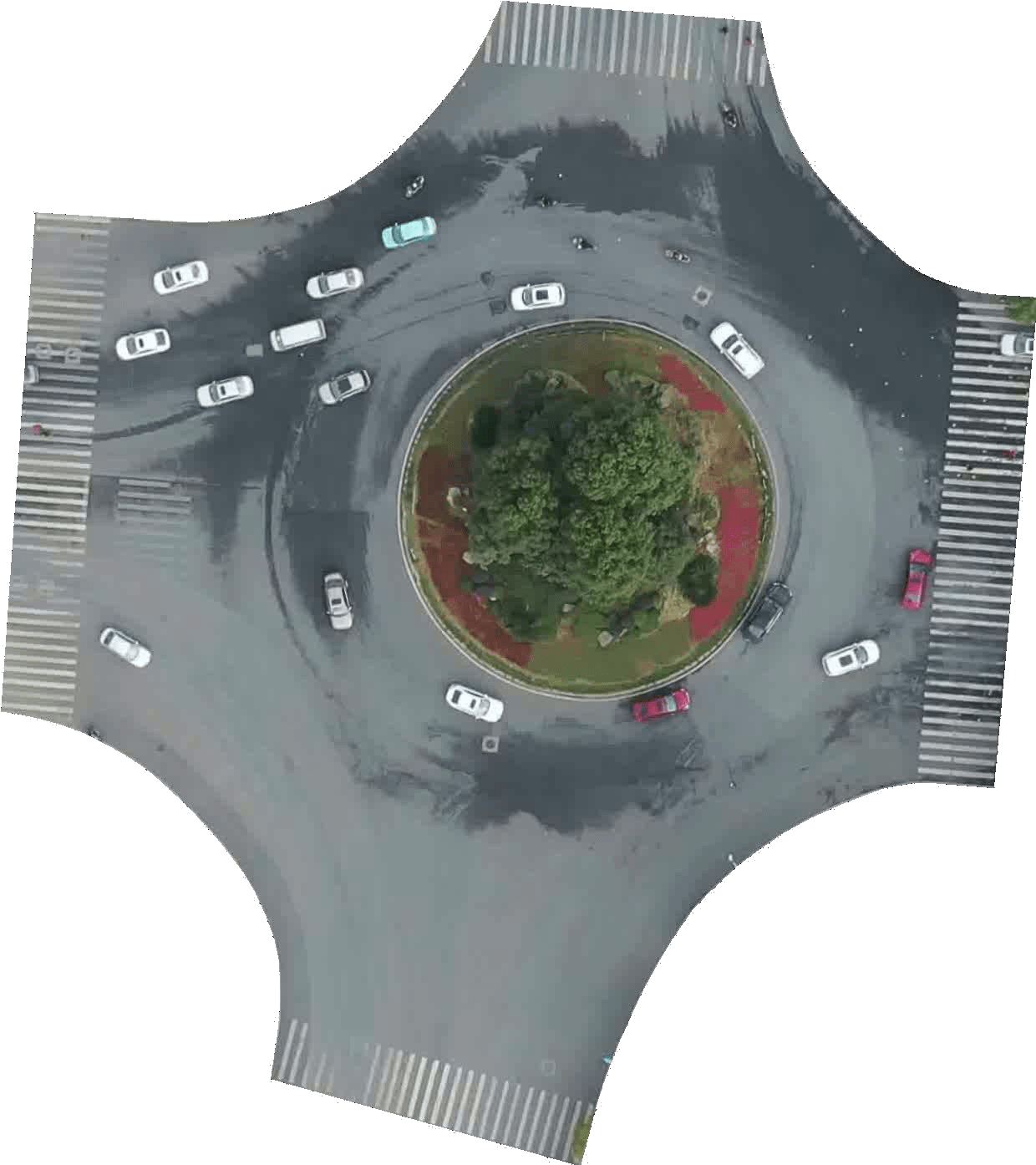}
      \caption{jnc06.mp4}
      \label{subfig:jnc06}
  \end{subfigure}
  \hfill
  \vspace{2em}
  \begin{subfigure}[t]{0.9\textwidth}
      \centering
      \includegraphics[width=\textwidth]{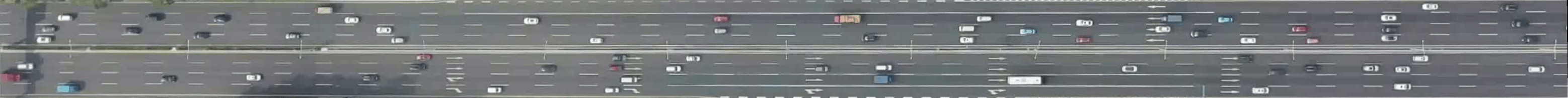}
      \caption{hwy00.mp4, hwy01.mp4, and hwy02.mp4}
      \label{subfig:hwy00}
  \end{subfigure}
  \hfill
  \begin{subfigure}[t]{\textwidth}
      \centering
      \includegraphics[width=\textwidth]{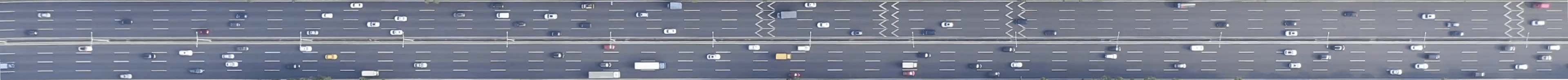}
      \caption{hwy03.mp4, hwy04.mp4, and hwy05.mp4}
      \label{subfig:hwy03}
  \end{subfigure}
  \hfill
  \begin{subfigure}[t]{0.8\textwidth}
      \centering
      \includegraphics[width=\textwidth]{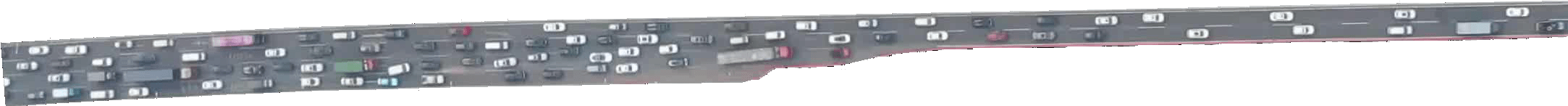}
      \caption{hwy06.mp4, hwy07.mp4, and hwy08.mp4}
      \label{subfig:hwy06}
  \end{subfigure}
  \hfill
  \begin{subfigure}[t]{0.45\textwidth}
      \centering
      \includegraphics[width=\textwidth]{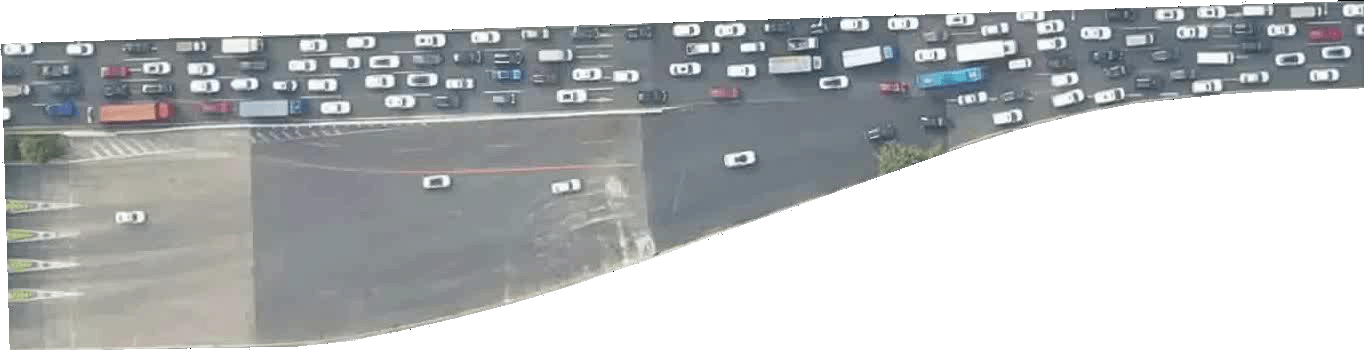}
      \caption{hwy09.mp4, hwy10.mp4, and hwy11.mp4}
      \label{subfig:hwy09}
  \end{subfigure}
  \vspace{2em}
  \caption{
   We propose a specialized dataset---Berkeley DeepDrive (B3D) dataset---and a modeling framework for studying decentralized vehicle coordination.
   The dataset, open sourced at \textbf{\color{cyan}https://github.com/b3d-project/b3d}, captures decentralized vehicle coordination on understructured roads.
   The modeling framework provide a novel perspective to distributed motion planning over networks.
  }
  \label{fig:topology}
  \end{figure}
}
\makeatother
\maketitle
\thispagestyle{empty}
\pagestyle{empty}

\begin{abstract}
  A significant portion of roads, particularly in densely populated developing countries, lacks explicitly defined right-of-way rules.
  These understructured roads pose substantial challenges for autonomous vehicle motion planning, where efficient and safe navigation relies on understanding decentralized human coordination for collision avoidance.
  This coordination, often termed ``social driving etiquette,'' remains underexplored due to limited open-source empirical data and suitable modeling frameworks.
  In this paper, we present a novel dataset and modeling framework designed to study motion planning in these understructured environments.
  The dataset includes 20 aerial videos of representative scenarios, an image dataset for training vehicle detection models, and a development kit for vehicle trajectory estimation.
  We demonstrate that a consensus-based modeling approach can effectively explain the emergence of priority orders observed in our dataset, and is therefore a viable framework for decentralized collision avoidance planning.
\end{abstract}

\section{Introduction}

Navigation in built environments---such as driving, cycling, and walking---is a crucial area of research in autonomous driving and human-robot interaction.
The development of autonomous agents capable of operating in \textit{structured} environments is a well-established field with roots extending back to the early days of control theory and robotics.
In the context of transportation, researchers have extensively studied navigation in structured environments, such as free-flow highways and signalized urban streets.
However, compared to navigation on structured roads, motion planning in \textit{understructured} road environments---roads without explicitly defined right-of-way regulations---is much less explored due to the lack of empirical data and the complexity of the problem.

The first essential ingredient for understanding understructured navigation is empirical data.
To this end, camera videos are particularly effective because \textsl{1}) they capture rich dynamics on roads at a relatively low cost, \textsl{2}) they allow for quantitative assessment through direct inspection, and \textsl{3}) they enables qualitative analysis through modern computer vision.
Despite extensive past research, most existing video datasets focus solely on driving behaviors in structured environments.
Behaviors in understructured road environments---such as crowded highways with frequent merges and unsignalized intersections---have rarely been surveyed or open sourced.
The scarcity of publicly available data for this problem undoubtedly hinders further investigation.

To bridge this gap in empirical data, we propose the Berkeley DeepDrive Drone (B3D) dataset.
This inertial-framed dataset records rich dynamics of driving behaviors in understructured road environments, including unsignalized intersections, unsignalized roundabouts, highways with collisions, highways with stop-and-go waves, and highways with merging bottlenecks, as shown in Figure~\ref{fig:topology}.
To the best of our knowledge, it is the first open-source drone dataset to date that extensively covers understructured driving behaviors.

The other piece of the puzzle is a suitable modeling paradigm.
The conventional control and planning architecture in an autonomous vehicle consists of four layers of abstraction, from top to bottom: \textsl{1}) routing, \textsl{2}) behavioral decision-making, \textsl{3}) motion planning, and \textsl{4}) vehicle control~\cite{paden2016survey}.
Particularly in the second behavioral layer, a prediction module is often employed to forecast the motion of surrounding vehicles, around which the motion planning layer then plans to avoid collisions.

This classical \textit{predict-then-plan} paradigm, while effective for driving in structured environments, is inadequate in understructured road environments.
For instance, at an unsignalized intersection, drivers negotiate the right-of-way dynamically: when two conflicting vehicles approach the intersection at similar times, the one perceived as more aggressive often ``wins'' the priority of passage.
This negotiative process fundamentally deviates from the predict-then-plan paradigm.

Modeling such a negotiative process is fundamental to autonomous driving.
Given the vast population of drivers in developing countries, where roads are often unsignalized, understanding how humans operate in these environments is crucial for enhancing the generalizability of autonomous driving technologies.
This knowledge enables autonomous vehicles to mimic human drivers, facilitating navigation through understructured road environments.
Furthermore, it may inspire innovative designs of decentralized motion planning algorithms, applicable not only to autonomous vehicles but also to other forms of mobile robots.

To this end, we propose a novel planning approach for the behavioral decision-making layer, replacing the traditional predict-then-plan approach with a \textit{negotiate-then-commit} approach.
Central to this new planning paradigm is a consensus model that implicitly arbitrates the priority of agents with conflicting paths.
This consensus model is founded on the principle of least action, a universal law observed in nature.

In summary, we list the main contributions of our work below.
\begin{itemize}
    \item We release the \textbf{B3D dataset}, which captures rich dynamics of driving behaviors on understructured roads.
    \item We propose a \textbf{consensus-based modeling framework} based on the principle of least action.
    \item We evaluate our framework through \textbf{data-driven validation and integrative simulation}.
\end{itemize}

\section{Related Works}
\label{sec:related_works}

Existing \textit{driving datasets} can be classified into either \textit{body-framed} dataset or \textit{inertial-framed} dataset.
The body-framed datasets have cameras placed on the traffic participants, e.g., on top a survey vehicle, to observe the movements of the surrounding traffic.
The inertial-framed datasets have cameras installed at some overhead positions above some roads of interest to observe all road occupants within its field of view.

Body-framed datasets are useful to study behaviors of the surrounding traffic participants with respect to the host.
It is arguably the most commonly used method for autonomous driving perception research.
Many well-known datasets fall within this category, such as the KITTI dataset~\cite{Geiger2013IJRR}, the Oxford RoboCar dataset~\cite{RobotCarDatasetIJRR}, the Cityscapes dataset~\cite{cordts2016cityscapes}, the Waymo Open dataset~\cite{sun2020scalability}, and the Waymo Open Motion dataset~\cite{ettinger2021large}.
Nevertheless, the body-framed placement makes it difficult to study persistent traffic patterns within a fixed spatial range.
Moreover, placing the sensors on a low-profile moving object also results in undesirable occlusion.

Inertial-framed datasets are better suited to observe traffic within a fixed spatial range.
By construction, the inertial-framed placement makes it very easy to estimate the recorded vehicle and human movements with respect to the ground.
Because the cameras are often placed on a vantage point, occlusion is also not as problematic.
In this category, we also find many seminal datasets, including the NGSIM dataset~\cite{ngsim2016}, the ARED dataset~\cite{wu2019tracking}, the Stanford Drone dataset~\cite{robicquet2020learning}, the highD dataset~\cite{highDdataset}, and the INTERACTION dataset~\cite{interactiondataset}.

Existing \textit{motion planning} literature for autonomous driving can be broadly divided into three major categories of methods: \textit{prediction-based} approaches, \textit{dynamics-based} approaches, and \textit{end-to-end} approaches.
Prediction-based planning deals with dynamically changing environments by predicting future states of the environment and planning based on those static predictions.
In contrast, dynamics-based planning maintains a model of the world and plans interactively based on that model.
Lastly, unlike the above two methods, end-to-end planning replaces conventional modular motion planning stack with a unified neural network model.

Prediction-based approaches are suitable for planning in environments where future states can be predicted with high confidence.
The predictability of such environments is characterized by the degree of confidence in forecasting the system's trajectory over a suitable time horizon.
For navigation in reasonably predictable environments, robust methods have been developed for planning around agents whose motions have bounded uncertainty~\cite{gray2013robust}, follow known probability distributions with chance constraints~\cite{luders2010chance,vitus2013probabilistic}, or are amenable to be described within a distributionally robust framework~\cite{renganathan2020towards,safaoui2024distributionally}.

Dynamics-based approaches are particularly suitable for highly coupled environments where the future state depends on the ego agent's current actions.
Within these approaches, consensus-based methods form a special class where all agents agree on a common navigation logic.
Such models are widely applied in driver and pedestrian behavior modeling~\cite{helbing1995social, treiber2000congested}, air traffic control~\cite{tomlin1998conflict, choi2024data}, and swarm robotics~\cite{reynolds1987flocks, trilaksono2015distributed}.
In contrast, non-consensus-based methods allow the ego agent and environmental agents to operate on different models, such as~\cite{sadigh2016planning}.

Recently, end-to-end approaches have gained significant attention in the community, driven by advances in machine learning, upgrade in computing infrastructure, and the availability of large-scale driving datasets.
For example, viable driving paths have been planned directly from LiDAR, GPS, IMU, and navigation map overlays~\cite{caltagirone2017lidar}, or from on-vehicle cameras and past vehicle states~\cite{xu2017end}.
The pros and cons of these approaches are clear: while they offer significant potential and scalability with large datasets and computing power, it lacks theoretical tractability and safety guarantees.
For a comprehensive overview of the literature in this area, please refer to~\cite{chen2024end} and the references therein.

\section{Berkeley DeepDrive Drone Dataset}
\label{sec:berkeley_deepdrive_drone_dataset}

To study decentralized vehicle coordination in understructured environments, we introduce the Berkeley DeepDrive Drone (B3D) dataset, available at \url{https://github.com/b3d-project/b3d}.
The dataset was recorded with a DJI Mavic 2 Pro quadcopter between December 11 and December 21 of 2019 in China.
It consists of 20 post-processed aerial drone videos, 16002 annotated images, and a development kit for estimating vehicle trajectories from the videos.
The total size of the dateset is about 86.3 GB.
We briefly describe the components of the dataset below.

\subsection{Aerial Videos}

Among the 20 processed aerial videos, eight were recorded on top of junctions and 12 on top of highways.
An overview of the types of the roads covered in the videos is illustrated in Figure~\ref{fig:topology}.
Scenarios recorded in the video can be classified into the following six categories:
\textsl{1}) unsignalized intersections,
\textsl{2}) unsignalized roundabouts,
\textsl{3}) tailgating accidents,
\textsl{4}) stop-and-go waves,
\textsl{5}) roadwork-induced merging, and
\textsl{6}) ramp-induced merging.

\textit{Unsignalized intersections} can be found in videos \lstinline{jnc00.mp4}, \lstinline{jnc01.mp4}, \lstinline{jnc02.mp4}, and \lstinline{jnc07.mp4}.
Videos \lstinline{jnc00.mp4} and \lstinline{jnc01.mp4} are two variants of three-way intersections, as shown in Figure~\ref{subfig:jnc00} and Figure~\ref{subfig:jnc01}, respectively.
Videos \lstinline{jnc02.mp4} and \lstinline{jnc07.mp4} are two variants of four-way intersections, as shown in Figure~\ref{subfig:jnc02} and Figure~\ref{subfig:jnc07}, respectively.

\textit{Unsignalized roundabouts} are captured in \lstinline{jnc03.mp4}, \lstinline{jnc04.mp4}, \lstinline{jnc05.mp4}, and \lstinline{jnc06.mp4}.
Videos \lstinline{jnc03.mp4} and \lstinline{jnc04.mp4} are two recordings of a five-way roundabout, as shown in Figure~\ref{subfig:jnc04}.
Videos \lstinline{jnc05.mp4} and \lstinline{jnc06.mp4} are two variants of four-way roundabouts, as shown in Figure~\ref{subfig:jnc05} and Figure~\ref{subfig:jnc06}, respectively.
Compared to video \lstinline{jnc03.mp4}, video \lstinline{jnc04.mp4} has slightly more traffic.

\begin{figure}[H]
  \centering
    \begin{subfigure}[t]{0.5\textwidth}
        \centering
        \begin{tikzpicture}[font=\sffamily\scriptsize]
          \node at (0,0)
            {\includegraphics[width=0.95\textwidth]{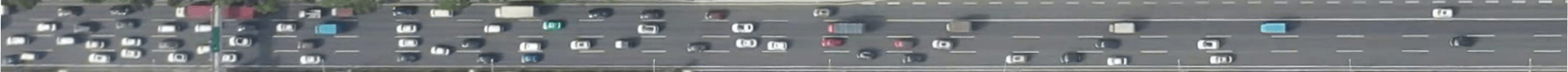}};
          \draw[very thick] (-4,-0.125) circle (4pt);
          \draw[] (-2,0.35) node[right] {First collision} -- (-4,0.35) -- (-4,0);
        \end{tikzpicture}
        \caption{First collision in \lstinline{hwy00.mp4}}
        \label{subfig:hwy00_collision}
    \end{subfigure}
    \hfill
    \begin{subfigure}[t]{0.5\textwidth}
        \centering
        \begin{tikzpicture}[font=\sffamily\scriptsize]
          \node at (0,0)
            {\includegraphics[width=0.95\textwidth]{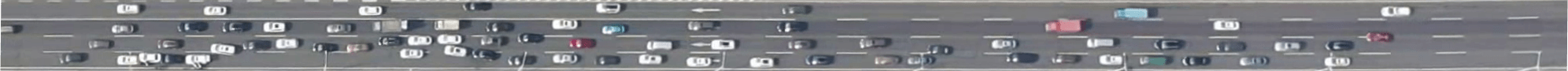}};
          \draw[very thick] (-3.25,-0.125) circle (4pt);
          \draw[] (-1.25,0.35) node[right] {Second collision} -- (-3.25,0.35) -- (-3.25,0);
        \end{tikzpicture}
        \caption{Second collision in \lstinline{hwy01.mp4}}
        \label{subfig:hwy01_collision}
    \end{subfigure}
  \caption{Tailgating collisions in \lstinline{hwy00.mp4} and \lstinline{hwy01.mp4}.
  The collided vehicles are circled in black.
  The first accident involves at least two vehicles, while the second incident involves four vehicles.}
  \label{fig:tailgate}
\end{figure}

The \textit{tailgating accidents} consist of two collision events, first in \lstinline{hwy00.mp4} and then in \lstinline{hwy01.mp4}.
At 00:45 of \lstinline{hwy00.mp4}, we observe the first accident near the left margin of the frame, as shown in Figure~\ref{subfig:hwy00_collision}.
At 13:10 of \lstinline{hwy01.mp4}, we find another traffic accident in the middle of the frame, as shown in Figure~\ref{subfig:hwy01_collision}.
Video \lstinline{hwy02.mp4} captures the resulting congested traffic induced by the second incident.
The the timestamps of the collision events are visualized in Figure~\ref{fig:timeline_tailgate}.

\begin{figure}[htbp]
  \centering
  \begin{tikzpicture}[%
      every node/.style={
          font=\sffamily\scriptsize,
          text height=1ex,
          text depth=.25ex,
      },
  ]
  \draw[->, >=latex] (0,0) -- (6.5,0);

  \foreach \x in {0,0.26,5.77}{
      \draw (\x cm,3pt) -- (\x cm,0pt);
  }

  \node[left] at (-0.05,0) {\textsl{hwy00.mp4}};
  \node[below] at (0,0) {0:00};
  \node[below] at (0.26,-0.2) {0:45};
  \draw[->] (0.26,-0.3) -- (0.26,0);
  \node[below] at (5.77,0) {16:31};

  \fill[myGreen] (0,0.25) rectangle (0.26,0.4);
  \fill[myRed] (0.26,0.25) rectangle (5.77,0.4);

  \draw[->] (0.26, 0.7) -- (0.26, 0.4);
  \node at (0.26,0.55) [right] {Collision 1};

  \end{tikzpicture}

  \vspace{-1.5em}
  \begin{tikzpicture}[%
      every node/.style={
          font=\sffamily\scriptsize,
          text height=1ex,
          text depth=.25ex,
      },
  ]
  \draw[->, >=latex] (0,0) -- (6.5,0);

  \foreach \x in {0,1.15,1.93,4.61,5.26}{
      \draw (\x cm,3pt) -- (\x cm,0pt);
  }

  \node[left] at (-0.05,0) {\textsl{hwy01.mp4}};
  \node[below] at (0,0) {0:00};
  \node[below] at (1.15,0) {3:18};
  \node[below] at (1.93,0) {5:30};
  \node[below] at (4.61,0) {13:10};
  \node[below] at (5.26,0) {15:01};

  \fill[myRed] (0,0.25) rectangle (1.15,0.4);
  \fill[myLightRed] (1.15,0.25) rectangle (1.93,0.4);
  \fill[myGreen] (1.93,0.25) rectangle (4.61,0.4);
  \fill[myRed] (4.61,0.25) rectangle (5.26,0.4);

  \draw[->] (1.15, 1) -- (1.15, 0.4);
  \node at (1.15,0.85) [right] {Collision 1 dissipating};
  \draw[->] (1.93, 0.7) -- (1.93, 0.4);
  \node at (1.93,0.55) [right] {Collision 1 dissipated};
  \draw[->] (4.61, 0.7) -- (4.61, 0.4);
  \node at (4.61,0.55) [right] {Collision 2};

  \end{tikzpicture}

  \vspace{-0.1em}
  \begin{tikzpicture}[%
      every node/.style={
          font=\sffamily\scriptsize,
          text height=1ex,
          text depth=.25ex,
      },
  ]
  \draw[->, >=latex] (0,0) -- (6.5,0);

  \foreach \x in {0,5.26}{
      \draw (\x cm,3pt) -- (\x cm,0pt);
  }

  \node[left] at (-0.05,0) {\textsl{hwy02.mp4}};
  \node[below] at (0,0) {0:00};
  \node[below] at (5.26,0) {15:01};

  \fill[myRed] (0,0.25) rectangle (5.26,0.4);

  \draw[->] (0, 0.7) -- (0, 0.4);
  \node at (0,0.55) [right] {Collision 2 continues};

  \end{tikzpicture}

  \caption{Timeline of the \textit{tailgating accidents}.
  Green indicates regular traffic.
  Red indicates congestion caused by a collision.
  Light red indicates the induced congestion starts dissipating.}
  \label{fig:timeline_tailgate}
\end{figure}

\textit{Stop-and-go waves} are recorded in \lstinline{hwy04.mp4} and \lstinline{hwy05.mp4}.
The first stop-and-go wave forms between 02:30 and 05:07 of \lstinline{hwy04.mp4}.
The second stop-and-go wave emerges between 06:06 and 08:10 of \lstinline{hwy04.mp4}.
The third stop-and-go wave is observed between 10:26 and 12:25 of \lstinline{hwy04.mp4}.
The fourth wave happens between 00:00 and 01:33 of \lstinline{hwy05.mp4}.
The last visible wave occurs between 05:19 and 06:07 of \lstinline{hwy05.mp4}.
For comparison, we provide video \lstinline{hwy03.mp4} as a free-flow baseline.
The formation and dissipation events of the stop-and-go waves are visualized in Figure~\ref{fig:timeline_wave}.

\begin{figure}[htbp]
  \centering
  \begin{tikzpicture}[%
      every node/.style={
          font=\sffamily\scriptsize,
          text height=1ex,
          text depth=.25ex,
      },
  ]
  \draw[->, >=latex] (0,0) -- (6.5,0);

  \foreach \x in {0,5.27}{
      \draw (\x cm,3pt) -- (\x cm,0pt);
  }

  \node[left] at (-0.05,0) {\textsl{hwy03.mp4}};
  \node[below] at (0,0) {0:00};
  \node[below] at (5.27,0) {15:03};

  \fill[myGreen] (0,0.25) rectangle (5.27,0.4);

  \draw[->] (0, 0.7) -- (0, 0.4);
  \node at (0,0.55) [right] {Freeflow};

  \end{tikzpicture}

  \vspace{-0.1em}
  \begin{tikzpicture}[%
      every node/.style={
          font=\sffamily\scriptsize,
          text height=1ex,
          text depth=.25ex,
      },
  ]
  \draw[->, >=latex] (0,0) -- (6.5,0);

  \foreach \x in {0,0.88,1.79,2.14,2.86,3.65,4.35,4.54,5.08}{
      \draw (\x cm,3pt) -- (\x cm,0pt);
  }

  \node[left] at (-0.05,0) {\textsl{hwy04.mp4}};
  \node[below] at (0,0) {0:00};
  \node[below] at (0.88,0) {2:30};
  \node[below] at (1.79,0) {5:07};
  \node[below] at (2.14,-0.2) {6:06};
  \draw[->] (2.14,-0.3) -- (2.14,0);
  \node[below] at (2.86,0) {8:10};
  \node[below] at (3.65,0) {10:26};
  \node[below] at (4.225,0) {12:25};
  \node[below] at (4.54,-0.2)  {12:58};
  \draw[->] (4.54,-0.3) -- (4.54,0);
  \node[below] at (5.08,0) {14:30};

  \fill[myGreen] (0,0.25) rectangle (0.88,0.4);
  \fill[myRed] (0.88,0.25) rectangle (1.79,0.4);
  \fill[myGreen] (1.79,0.25) rectangle (2.14,0.4);
  \fill[myRed] (2.14,0.25) rectangle (2.86,0.4);
  \fill[myGreen] (2.86,0.25) rectangle (3.65,0.4);
  \fill[myLightRed] (3.65,0.25) rectangle (4.35,0.4);
  \fill[myGreen] (4.35,0.25) rectangle (4.54,0.4);
  \fill[myRed] (4.54,0.25) rectangle (5.08,0.4);

  \draw[->] (0.88, 0.7) -- (0.88, 0.4);
  \node at (0.88,0.55) [right] {Wave 1};
  \draw[->] (2.14, 0.7) -- (2.14, 0.4);
  \node at (2.14,0.55) [right] {Wave 2};
  \draw[->] (3.65, 0.7) -- (3.65, 0.4);
  \node at (3.6,0.55) [right] {Wave 3};
  \draw[->] (4.54, 0.7) -- (4.54, 0.4);
  \node at (4.54,0.55) [right] {Wave 4};

  \end{tikzpicture}

  \vspace{-0.1em}
  \begin{tikzpicture}[%
      every node/.style={
          font=\sffamily\scriptsize,
          text height=1ex,
          text depth=.25ex,
      },
  ]
  \draw[->, >=latex] (0,0) -- (6.5,0);

  \foreach \x in {0,0.54,1.86,2.14,5.27}{
      \draw (\x cm,3pt) -- (\x cm,0pt);
  }

  \node[left] at (-0.05,0) {\textsl{hwy05.mp4}};
  \node[below] at (0,0) {0:00};
  \node[below] at (0.54,0) {1:33};
  \node[below] at (1.86,0) {5:19};
  \node[below] at (2.14,-0.2) {6:07};
  \draw[->] (2.14,-0.3) -- (2.14,0);
  \node[below] at (5.27,0) {15:04};

  \fill[myRed] (0,0.25) rectangle (0.54,0.4);
  \fill[myGreen] (0.54,0.25) rectangle (1.86,0.4);
  \fill[myLightRed] (1.86,0.25) rectangle (2.14,0.4);
  \fill[myGreen] (2.14,0.25) rectangle (5.27,0.4);

  \draw[->] (0, 0.7) -- (0, 0.4);
  \node at (0,0.55) [right] {Wave 5};
  \draw[->] (1.86, 0.7) -- (1.86, 0.4);
  \node at (1.86,0.55) [right] {Wave 6};

  \end{tikzpicture}

  \caption{Timeline of the \textit{stop-and-go waves}.
  Green indicates regular traffic.
  Red indicates congestion caused by a strong stop-and-go wave.
  Light red indicates congestion caused by a weak stop-and-go wave.}
  \label{fig:timeline_wave}
\end{figure}

\textit{Roadwork-induced merging} is recorded in \lstinline{hwy06.mp4}, \lstinline{hwy07.mp4}, and \lstinline{hwy08.mp4}.
The topology of the scenario is a four-lane-to-two-lane bottleneck, as shown in Figure~\ref{subfig:hwy06}.
Persistent congestion is observed before the merge point, while free-flow traffic is formed after the merge point.

\textit{Ramp-induced merging} is recorded in \lstinline{hwy09.mp4}, \lstinline{hwy10.mp4}, and \lstinline{hwy11.mp4}.
The topology of the ramp is shown in Figure~\ref{subfig:hwy09}, where a three-lane on-ramp is merging into a four-lane congested highway.
The traffic stays congested \textit{before and after} the merge point.

\subsection{Annotated Images}

In addition to the videos, we build an image dataset, which can be used to train vehicle detection models for trajectory estimation.
The dataset consists of 16002 annotated images, 80\% of which is split for training, 10\% of which for validation, and 10\% of which for testing.

Using annotation tool CVAT~\cite{boris_sekachev_2020_4009388}, a total of 135303 axially aligned bounding boxes (AABB) are annotated for junction images and a total of 129939 AABBs created for highway images.
Note that the annotations contain only one object class, that is, the \textit{vehicle} class.
We do not distinguish large vehicles, such as buses and trucks, from small vehicles, such as sedans and SUVs.

\subsection{Development Kit}

Besides, we also provide a development kit consisting of three example scripts: train.py, test.py, and mask.py.
For reproducibility, we encapsulate the development environment in a Docker image.
The image can be directly built from the provided Dockerfile in the dataset repository.

The script train.py is used to show how the annotated image data may be used to train a neural network model for vehicle detection.
In this script, we use the object detection library Detectron2~\cite{wu2019detectron2} to train a RetinaNet model~\cite{lin2017focal} for detecting locations of vehicles in an input image.

The script test.py applies the trained model, trained via the train.py script, to an input image.
This script is intended to serve as an example for post-training evaluation and inference.
For convenience, we provide an optional pre-trained model, which can be used directly for inference.
This way, users can directly use a working detection model to estimate vehicle trajectories without having to go through the computationally expensive training step locally.

Finally, the script mask.py crops an image according to a pre-defined polygon mask.
The script intends to help users to focus on the only relevant part of the scene, where relevance is defined by the user via CVAT.
To crop a video, one only needs to define a polygonal mask for \textit{one} frame of a video and then apply masking to \textit{every} frame of the video.

\section{Consensus-Based Models}
\label{sec:consensus_based_models}

To describe the decentralized vehicle coordination observed in the B3D dataset, we propose a modeling framework based on the idea of consensus.

We consider a setup with $N$ agents and $M$ pairs of intersecting paths.
Denote the road network by $G = (V, E)$, where $V$ represents the set of vertices and $E$ the set of edges.
A path $p$ is defined as a finite sequence of directed edges connecting a set of unique vertices.

Let $s \in \mathbb{R}_{\geq 0}$ represent the traveled distance along path $p$.
Define the state of an agent as $\vec{x} \coloneqq [s \quad \dot{s}]^{\top}$.
Assuming the agent begins its path at $t_{0}$ and completes it at $t_{f}$, a \textit{traversal policy} $\pi : \mathbb{R} \times \mathbb{R}^{2} \rightarrow \mathbb{R}^{2}$ is a function defined as $\dot{\vec{x}}(t) = \pi(t, \vec{x}(t))$ over $t \in [t_{0}, t_{f}]$.
The trajectory of an agent is defined as the evolution of an agent's state over time, i.e., $X = \{\vec{x}(t) \mid \forall t \in [t_{0}, t_{f}]\}$.

Two traversal policies $\pi_{1}: [t_{0}, t_{f}] \rightarrow \mathbb{R}^{2}$ and $\pi_{2}: [t_{0}, t_{f}] \rightarrow \mathbb{R}^{2}$ are considered collision-free if the resulting trajectories $\vec{x}_{1}(t)$ and $\vec{x}_{2}(t)$ remain sufficiently separated according to some distance function $d: \mathbb{R}^{2} \times \mathbb{R}^{2} \rightarrow \mathbb{R}_{\geq 0}$ for all $t \in [t_{0}, t_{f}]$.
Specifically, we require that $d(\vec{x}_{1}(t), \vec{x}_{2}(t)) \geq \epsilon$ for a certain separation margin $\epsilon > 0$ and for all $t$.


With the preceding definition, we are now equipped to present the two-agent collision avoidance problem as follows:
\begin{problem}{(Two-Agent Collision Avoidance Problem)}
    \label{prob:two_agent}
    Consider two agents of paths $p_1$ and $p_2$ intersecting at either one vertice or a sub-path.
    Given the initial conditions $(t_{0}, \vec{x}_{0 \mid 1})$ and $(t_{0}, \vec{x}_{0 \mid 2})$, find a pair of traversal policies, $\pi_1$ and $\pi_2$, such that they establish a priority order at the conflicting location with finite arrival times $t_{f \mid 1}$ and $t_{f \mid 2}$.
\end{problem}

The above two-agent problem is an special case of the more general $N$-agent problem, as formally stated below:
\begin{problem}{($N$-Agent Collision Avoidance Problem)}
    \label{prob:n_agent}
    Consider $N$ agents of paths $P = \{p_{1}, p_{2}, \dots, p_{N}\}$ intersecting at either one vertice or a sub-path.
    Given the initial conditions $(t_{0}, \vec{x}_{0 \mid i}), \forall i = 1, \dots, N$, find some traversal policies $\pi_{1}, \pi_{2}, \dots, \pi_{N}$, such that they solve the corresponding two-agent collision avoidance problem for each unique pair of paths in $P$.
\end{problem}

\subsection{Collision Avoidance Consensus}
A consensus model, informally, refers to a control policy that, when deploy to individual agents, steers each agent towards consensus on certain critical quantities despite initial disagreements.
For Problem~\ref{prob:two_agent} and Problem~\ref{prob:n_agent}, the critical quantity of interest here is the priority order at all conflicting locations.
In following paragraphs, we present a general framework on consensus-based conflict resolution.

Without loss of generality, we start by first considering Problem~\ref{prob:two_agent}.
Label the two agents as Agent 1 and Agent 2.
Define a binary indicator $r_{i} \in \{0, 1\}$ at the conflicting location for agent $i \in \{1, 2\}$ such that $r_{i} = 1$ when Agent $i$ has the priority and $r_{i} = 0$ otherwise.
For the two-agent problem, a priority order is established if and only if $r_{1} + r_{2} = 1$.

Consequently, we assume a \textit{consensus model} for collision avoidance $\sigma: \mathbb{R} \times \mathbb{R}^{2} \times \mathbb{R}^{2} \rightarrow \{0, 1\}$ to take the following form:
\begin{equation}
    r = \sigma(t, \vec{x}(t), \bar{\vec{x}}(t)),
\end{equation}
where $t$ is the time of evaluation and $\vec{x}(t), \bar{\vec{x}}(t)$ the states of the ego agent and the competing agent, respectively.
Clearly, a valid consensus model satisfies $\sigma(t, \vec{x}_{1}(t), \vec{x}_{2}(t)) + \sigma(t, \vec{x}_{2}(t), \vec{x}_{1}(t)) = 1$, where $\vec{x}_{1}(t), \vec{x}_{2}(t)$ are the states of Agent 1 and Agent 2.


Now consider the general $N$-agent version in Problem~\ref{prob:n_agent}.
By similar argument, we must have $\sigma(t, \vec{x}_{p}(t), \vec{x}_{q}(t)) + \sigma(t, \vec{x}_{q}(t), \vec{x}_{p}(t)) = 1$, where $\vec{x}_{p}(t), \vec{x}_{q}(t)$ are the states of any unique pair of Agent $p$ and Agent $q$ selected from the $N$ agents.

Altternatively, we can define $r_{i} \in \{0, 1, \dots, N-1\}$ at the conflicting location for Agent $i \in \{1, \dots, N\}$ such that the set $R = \{r_{i} : \forall i = 1, \dots, N\}$ is a permutation of the ordered set $\{0, 1, \dots, N-1\}$.
The priority in which agent $i$ clears the conflicting location is denoted by $r_{i}$, i.e., if $r_{i} > r_{j}$, then Agent $i$ clears the conflicting location \textit{earlier} than Agent $j$.

Next, we propose a specific form of consensus model based on the principle of least action.

\subsection{Least-Action Consensus}


To develop a least-action consensus model, one must first quantify the cost of an action associated with each conflict resolution scheme.
To this end, a common approach is to measure the cost of an action through a cost functional $J$ on the trajectories resulting from a given priority order.

For example, we can define a $L^{2}$-norm induced cost functional.
Given a priority order $R = \{r_{i} : \forall i = 1, \dots, N\}$ and initial conditions $I = \{(t_{0}, \vec{x}_{0 \mid i}) : \forall i = 1, \dots, N\}$ for the $N$ agents respectively, we can find a set of valid traversal policies $\Pi_{R,I} = \{\pi_{i} : \forall i = 1, \dots, N\}$.
Let $T_{R,I}^{*} = \{[t_{i}^{-}, t_{i}^{+}] : i = 1, \dots, N\}$ be the set of time intervals, during which the respective agents interact with their neighboring agents.
With $R, I, T_{R,I}^{*}$, one can therefore define the cost functional $J$ as follows:
\begin{equation}
    J(R, I, \Pi_{R,I}, T_{R,I}^{*}) = \sum_{i=1}^{N} \left(\int_{t_{i}^{-}}^{t_{i}^{+}} \pi_{i}^{2}(t, \vec{x}_{i}(t)) \, dt \right)^{\frac{1}{2}},
\end{equation}
for $[t_{i}^{-}, t_{i}^{+}] \in T_{R,I}^{*}, \pi_{i} \in \Pi_{R, I}$ for all $i = 1, \dots, N$.

In Problem~\ref{prob:two_agent}, if one can associate a cost functional with each valid choice of traversal policies $\pi_{1}, \pi_{2}$ and initial conditions $(t_{0}, \vec{x}_{0 \mid 1})$ and $(t_{0}, \vec{x}_{0 \mid 2})$, then a least-action consensus model favors the choice associated with the least cost.
With the cost functional $J$, this leads to solving the following optimization problem:
\begin{argmini}
  {R}{J(R, I, \Pi_{R,I}, T_{R,I}^{*})}{\label{eq:two_agent}}{R^{*} = }
  \addConstraint{r_{1} + r_{2}}{=1}
  \addConstraint{r_{1}, r_{2}}{\in \{0, 1\}}.
\end{argmini}
Let $\sigma(t_{0}, \vec{x}_{0 \mid 1}, \vec{x}_{0 \mid 2}) \coloneqq r_{1}^{*}$ and $\sigma(t_{0}, \vec{x}_{0 \mid 2}, \vec{x}_{0 \mid 1}) \coloneqq r_{2}^{*}$.
Clearly, if the optimization~\eqref{eq:two_agent} admits an unique solution, the resulting consensus model ensures zero collision.

Generalizing to Problem~\ref{prob:n_agent}, we find the total number of possible orderings extends to $N!$.
Using the same cost functional, the consensus model enumerates the costs for each possible orderings and select the one with the least amount of costs.
, as defined below:
\begin{argmini}
  {R}{J(R, I, \Pi_{R,I}, T_{R,I}^{*})}{\label{eq:n_agent}}{R^{*} = }
  \addConstraint{R}{\in S(\{0, 1, \dots, N-1\})},
\end{argmini}
where $S(\{0, 1, \dots, N-1\})$ is the set of all permuations of $\{0, 1, \dots, N-1\}$.
Let $\sigma(t_{0}, \vec{x}_{0 \mid i}, \vec{x}_{0 \mid j}) \coloneqq \mathds{1}(r_{i} > r_{j})$ for all $i = 1, \dots, N$.
As expected, optimization~\eqref{eq:two_agent} is a special case of optimization~\eqref{eq:n_agent}, where $N = 2$.
Like before, if the optimization~\eqref{eq:n_agent} admits an unique solution, the resulting consensus leads to zero collision.

\begin{remark}
  Because collision avoidance is inherently a local phenomenon, explicit communication of the coupling agents' states is unnecessary.
  Instead, state awareness can be achieved through local sensing.
  Furthermore, explicit action sharing is also not required, as local execution of the consensus model guarantees conflict resolution.
\end{remark}

\begin{remark}
  The time interval $[t_{i}^{-}, t_{i}^{+}]$ marks the begining and the end of Agent $i$'s interaction with other conflicting agents.
  When setting the time interval, one can choose $t_{i}^{-}$ as a few seconds before entering the conflicting location and $t_{i}^{+}$ as the time of exiting the conflicting location.
\end{remark}

\begin{remark}
  Cost disparity among options can indicate confidence level.
  A greater cost difference strengthens consensus.
  Moreover, a slack variable $\xi$ can model agent defensiveness or aggressiveness.
  For example, for plans with costs $J_{1}$ and $J_{2}$, we can introduce $\xi$ such that plan 1 is preferred if $J_{1} < J_{2} + \xi$, and vice versa.
  Negative $\xi$ represents a defensive agent, while positive $\xi$ indicates an aggressive agent.
\end{remark}

\begin{remark}
In the events where optimization~\eqref{eq:two_agent} or~\eqref{eq:n_agent} does not admit an unique solution, it becomes necessary to implement additional logic for tie-breaking.
In pactice, this may entail that all conflicting vehicles simultaneously coming to a stop, thereafter proceeding based on established social norms or sign language cues.
\end{remark}

\section{Numerical Simulation}
\label{sec:numerical_simulation}

In this section, we evaluate our least-action consensus model by conducting tests using \textsl{1}) an sequence of two-agent validation scenarios with ground truth provided by the B3D dataset and \textsl{2}) an integrative $N$-agent simulation scenario.

\subsection{Two-Agent Validation}

The two-agent scenarios are extracted from a specific unsignalized intersection as recorded in video ``jnc07.mp4'' shown in Figure~\ref{subfig:jnc07}.
The primary objective is to determine whether our least-action consensus model recovers the order of proceeding observed in the video.
We manually selected six relatively isolated two-agent interactions from this video to apply the least-action consensus model.

Results from the two-agent validation tests are presented in Table~\ref{tab:two_agent}.
As indicated in the table, the model successfully recovers naturalistic conflict resolution under all initial conditions.

\begin{table}[htbp]
    \centering
    \caption{Results of the Two-Agent Validation}
    \resizebox{0.49\textwidth}{!}{\begin{tabular}{cccc}
    \hline
    Entry Speeds (m$/$s)* & Distances to Conflict (m)* & True Priority Order & Modelled Priority Order \\
    \hline
    (0.01, 3.57)       & (5.52, 21.25)             & Westbound First     & Westbound First         \\
    (0.01, 5.32)       & (5.37, 20.04)             & Westbound First     & Westbound First         \\
    (0.02, 3.24)       & (5.87, 21.14)             & Westbound First     & Westbound First         \\
    (0.16, 2.65)       & (5.82, 21.71)             & Westbound First     & Westbound First         \\
    (0.04, 1.79)       & (7.33, 21.97)             & Westbound First     & Westbound First         \\
    (0.41, 3.18)       & (4.35, 21.19)             & Southbound First    & Southbound First\\
    \hline
    \multicolumn{4}{l}{\small * Data on entry speeds and distance to conflict are arranged as (southbound, westbound)} \\
    \end{tabular}}
    \label{tab:two_agent}
\end{table}

\subsection{$N$-Agent Simulation}

\begin{figure}[htbp]
    \centering
    \includegraphics[width=0.175\textwidth]{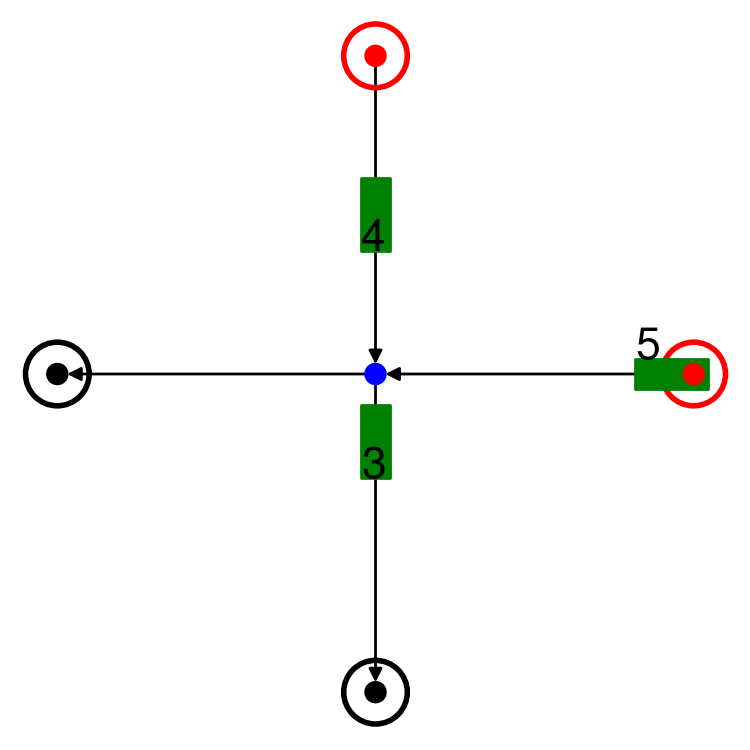}
    \caption{The $N$-agent simulation scenario.
    Red circles: trip origins.
    Black circles: destinations.
    Numbered green rectangles: vehicles at the intersection.
    }
    \label{fig:n_agent_scenario}
\end{figure}

To evaluate the integrative performance of the model, we constructed a $N$-agent cross-intersection simulation involving southbound and westbound traffic flows as shown in Figure~\ref{fig:n_agent_scenario}.
The goal of this evaluation is to check if our least-action model can be deployed in an integrative simulation without collision.
Each arm of the cross-intersection is symmetrically configured to be 20 meters in length.
The simulation is designed to run for a total duration of approximately 300 seconds, during which inbound traffic is randomly generated based on a Poisson distribution with an average arrival rate of 900 vehicles per hour.
The initial position of each vehicle is set at the respective intersection entrances, and the initial velocity is sampled from a normal distribution with a mean of 3 m$/$s and a standard deviation of 1 m$/$s, truncated between 0 and 4 m$/$s.

\begin{figure}[htbp]
    \centering
    \includegraphics[width=0.5\textwidth]{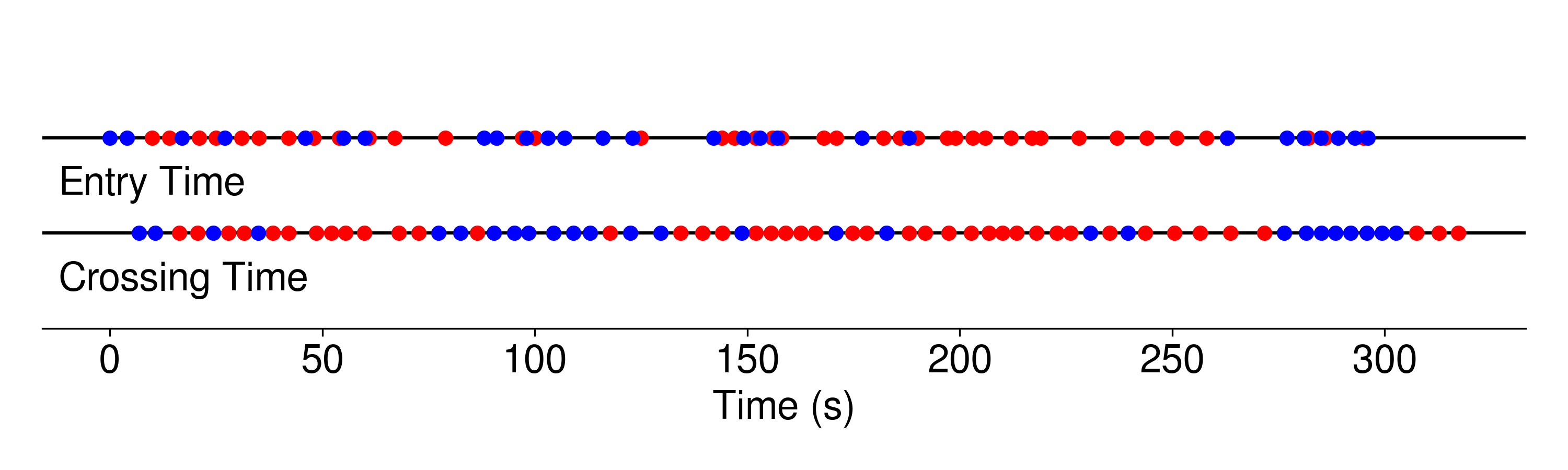}
    \caption{Results of the $N$-agent simulation.
    Red points: southbound vehicles.
    Blue points: westbound vehicles.
    }
    \label{fig:gantt_chart}
\end{figure}

The results of the $N$-agent simulation tests are depicted in Figure~\ref{fig:gantt_chart}.
The figure features two timelines: the top timeline visualizes the entry times of incoming vehicles, while the bottom timeline displays their crossing times.
Red data points represent vehicles approaching from the southbound direction, and blue data points from the westbound.
Although vehicles enter the intersection randomly, as per the simulation setup, they cross the conflict point in a distinct batch pattern.
Each direction alternates in crossing the conflict point in batches of queues.
This qualitative behavior aligns with observations from the B3D dataset.

\section{Discussion}
\label{sec:discussion}

The preliminary results from the numerical simulations suggest that the consensus-based approach is a viable framework for decentralized conflict resolution in understructured road environments.
Despite these promising findings, several critical aspects of this research remain unexplored.

A key theoretical challenge is demonstrating consensus convergence under disturbances.
While empirical results indicate this approach reflects human driver behavior in real-time negotiation, conditions for guaranteed convergence to a consistent priority order in distributed settings has yet to be proposed and proven.

Further empirical evaluation of model performance is crucial, especially with complex testing. For example, testing should expand beyond two-agent conflicts to $N$-agent interactions with diverse conflict types and demand profiles.  Furthermore, Turing-test-like evaluations of the consensus model are of interest; a successful model should be indistinguishable from human drivers to an observer in understructured environments.

Finally, modeling the ``handshake phase''---how vehicles become locally coupled as they approach imminent collisions---remains an open question.
Clearly, coupling mechanism during this phase is influenced by vehicle paths, proximity, speeds, and driver characteristics.
Developing a succinct model capturing essential features while disregarding less relevant factors presents a significant challenge.






\bibliographystyle{plain}
\bibliography{references}

\begin{thebibliography}{10}

\bibitem{ngsim2016}
United States Federal~Highway Administration.
\newblock Next generation simulation (ngsim) vehicle trajectories and supporting data, 2016.

\bibitem{caltagirone2017lidar}
Luca Caltagirone, Mauro Bellone, Lennart Svensson, and Mattias Wahde.
\newblock Lidar-based driving path generation using fully convolutional neural networks.
\newblock In {\em 2017 IEEE 20th International Conference on Intelligent Transportation Systems (ITSC)}, pages 1--6. IEEE, 2017.

\bibitem{chen2024end}
Li~Chen, Penghao Wu, Kashyap Chitta, Bernhard Jaeger, Andreas Geiger, and Hongyang Li.
\newblock End-to-end autonomous driving: Challenges and frontiers.
\newblock {\em IEEE Transactions on Pattern Analysis and Machine Intelligence}, 2024.

\bibitem{choi2024data}
Hong-Cheol Choi and Inseok Hwang.
\newblock Data-driven trajectory-based consensus approach to traffic management for manned and unmanned aviation.
\newblock In {\em AIAA SCITECH 2024 Forum}, page 0537, 2024.

\bibitem{cordts2016cityscapes}
Marius Cordts, Mohamed Omran, Sebastian Ramos, Timo Rehfeld, Markus Enzweiler, Rodrigo Benenson, Uwe Franke, Stefan Roth, and Bernt Schiele.
\newblock The cityscapes dataset for semantic urban scene understanding.
\newblock In {\em Proceedings of the IEEE conference on computer vision and pattern recognition}, pages 3213--3223, 2016.

\bibitem{ettinger2021large}
Scott Ettinger, Shuyang Cheng, Benjamin Caine, Chenxi Liu, Hang Zhao, Sabeek Pradhan, Yuning Chai, Ben Sapp, Charles Qi, Yin Zhou, Zoey Yang, Aurelien Chouard, Pei Sun, Jiquan Ngiam, Vijay Vasudevan, Alexander McCauley, Jonathon Shlens, and Dragomir Anguelov.
\newblock Large scale interactive motion forecasting for autonomous driving: The waymo open motion dataset.
\newblock In {\em Proceedings of the IEEE/CVF International Conference on Computer Vision}, pages 9710--9719, 2021.

\bibitem{Geiger2013IJRR}
Andreas Geiger, Philip Lenz, Christoph Stiller, and Raquel Urtasun.
\newblock Vision meets robotics: The kitti dataset.
\newblock {\em International Journal of Robotics Research}, 32(11):1231--1237, 2013.

\bibitem{gray2013robust}
Andrew Gray, Yiqi Gao, J~Karl Hedrick, and Francesco Borrelli.
\newblock Robust predictive control for semi-autonomous vehicles with an uncertain driver model.
\newblock In {\em 2013 IEEE intelligent vehicles symposium (IV)}, pages 208--213. IEEE, 2013.

\bibitem{helbing1995social}
Dirk Helbing and Peter Molnar.
\newblock Social force model for pedestrian dynamics.
\newblock {\em Physical review E}, 51(5):4282, 1995.

\bibitem{highDdataset}
Robert Krajewski, Julian Bock, Laurent Kloeker, and Lutz Eckstein.
\newblock The highd dataset: A drone dataset of naturalistic vehicle trajectories on german highways for validation of highly automated driving systems.
\newblock In {\em International Conference on Intelligent Transportation Systems}, pages 2118--2125, 2018.

\bibitem{lin2017focal}
Tsung-Yi Lin, Priya Goyal, Ross Girshick, Kaiming He, and Piotr Doll{\'a}r.
\newblock Focal loss for dense object detection.
\newblock In {\em Proceedings of the IEEE International Conference on Computer Vision}, pages 2980--2988, 2017.

\bibitem{luders2010chance}
Brandon Luders, Mangal Kothari, and Jonathan How.
\newblock Chance constrained rrt for probabilistic robustness to environmental uncertainty.
\newblock In {\em AIAA guidance, navigation, and control conference}, page 8160, 2010.

\bibitem{RobotCarDatasetIJRR}
Will Maddern, Geoff Pascoe, Chris Linegar, and Paul Newman.
\newblock 1 year, 1000km: The oxford robotcar dataset.
\newblock {\em The International Journal of Robotics Research}, 36(1):3--15, 2017.

\bibitem{paden2016survey}
Brian Paden, Michal {\v{C}}{\'a}p, Sze~Zheng Yong, Dmitry Yershov, and Emilio Frazzoli.
\newblock A survey of motion planning and control techniques for self-driving urban vehicles.
\newblock {\em IEEE Transactions on intelligent vehicles}, 1(1):33--55, 2016.

\bibitem{renganathan2020towards}
Venkatraman Renganathan, Iman Shames, and Tyler~H Summers.
\newblock Towards integrated perception and motion planning with distributionally robust risk constraints.
\newblock {\em IFAC-PapersOnLine}, 53(2):15530--15536, 2020.

\bibitem{reynolds1987flocks}
Craig~W Reynolds.
\newblock Flocks, herds and schools: A distributed behavioral model.
\newblock In {\em Proceedings of the 14th annual conference on Computer graphics and interactive techniques}, pages 25--34, 1987.

\bibitem{robicquet2020learning}
A~Robicquet, A~Sadeghian, A~Alahi, and S~Savarese.
\newblock Learning social etiquette: Human trajectory prediction in crowded scenes.
\newblock In {\em European Conference on Computer Vision}, volume~2, 2020.

\bibitem{sadigh2016planning}
Dorsa Sadigh, Shankar Sastry, Sanjit~A Seshia, and Anca~D Dragan.
\newblock Planning for autonomous cars that leverage effects on human actions.
\newblock In {\em Robotics: Science and systems}, volume~2, pages 1--9. Ann Arbor, MI, USA, 2016.

\bibitem{safaoui2024distributionally}
Sleiman Safaoui and Tyler~H Summers.
\newblock Distributionally robust cvar-based safety filtering for motion planning in uncertain environments.
\newblock In {\em 2024 IEEE International Conference on Robotics and Automation (ICRA)}, pages 103--109. IEEE, 2024.

\bibitem{boris_sekachev_2020_4009388}
Boris Sekachev, Nikita Manovich, Maxim Zhiltsov, Andrey Zhavoronkov, Dmitry Kalinin, Ben Hoff, TOsmanov, Dmitry Kruchinin, Artyom Zankevich, DmitriySidnev, Maksim Markelov, Johannes222, Mathis Chenuet, a~andre, telenachos, Aleksandr Melnikov, Jijoong Kim, Liron Ilouz, Nikita Glazov, Priya4607, Rush Tehrani, Seungwon Jeong, Vladimir Skubriev, Sebastian Yonekura, vugia truong, zliang7, lizhming, and Tritin Truong.
\newblock opencv/cvat: v1.1.0, 8 2020.

\bibitem{sun2020scalability}
Pei Sun, Henrik Kretzschmar, Xerxes Dotiwalla, Aurelien Chouard, Vijaysai Patnaik, Paul Tsui, James Guo, Yin Zhou, Yuning Chai, Benjamin Caine, Vijay Vasudevan, Wei Han, Jiquan Ngiam, Hang Zhao, Aleksei Timofeev, Scott Ettinger, Maxim Krivokon, Amy Gao, Aditya Joshi, Sheng Zhao, Shuyang Cheng, Yu~Zhang, Jonathon Shlens, Zhifeng Chen, and Dragomir Anguelov.
\newblock Scalability in perception for autonomous driving: Waymo open dataset.
\newblock In {\em Proceedings of the IEEE/CVF conference on computer vision and pattern recognition}, pages 2446--2454, 2020.

\bibitem{tomlin1998conflict}
Claire Tomlin, George~J Pappas, and Shankar Sastry.
\newblock Conflict resolution for air traffic management: A study in multiagent hybrid systems.
\newblock {\em IEEE Transactions on automatic control}, 43(4):509--521, 1998.

\bibitem{treiber2000congested}
Martin Treiber, Ansgar Hennecke, and Dirk Helbing.
\newblock Congested traffic states in empirical observations and microscopic simulations.
\newblock {\em Physical Review E}, 62(2):1805, 2000.

\bibitem{trilaksono2015distributed}
Bambang~Riyanto Trilaksono.
\newblock Distributed consensus control of robot swarm with obstacle and collision avoidance.
\newblock In {\em 2015 2nd International Conference on Information Technology, Computer, and Electrical Engineering (ICITACEE)}, pages 2--2. IEEE, 2015.

\bibitem{vitus2013probabilistic}
Michael~P Vitus and Claire~J Tomlin.
\newblock A probabilistic approach to planning and control in autonomous urban driving.
\newblock In {\em 52nd IEEE Conference on Decision and Control}, pages 2459--2464. IEEE, 2013.

\bibitem{wu2019tracking}
Fangyu Wu, Raphael~E Stern, Shumo Cui, Maria~Laura Delle~Monache, Rahul Bhadani, Matt Bunting, Miles Churchill, Nathaniel Hamilton, Benedetto Piccoli, Benjamin Seibold, Jonathan Sprinkle, and Daniel Work.
\newblock Tracking vehicle trajectories and fuel rates in phantom traffic jams: Methodology and data.
\newblock {\em Transportation Research Part C: Emerging Technologies}, 99:82--109, 2019.

\bibitem{wu2019detectron2}
Yuxin Wu, Alexander Kirillov, Francisco Massa, Wan-Yen Lo, and Ross Girshick.
\newblock Detectron2, 2019.

\bibitem{xu2017end}
Huazhe Xu, Yang Gao, Fisher Yu, and Trevor Darrell.
\newblock End-to-end learning of driving models from large-scale video datasets.
\newblock In {\em Proceedings of the IEEE conference on computer vision and pattern recognition}, pages 2174--2182, 2017.

\bibitem{interactiondataset}
Wei Zhan, Liting Sun, Di~Wang, Haojie Shi, Aubrey Clausse, Maximilian Naumann, Julius Kummerle, Hendrik Konigshof, Christoph Stiller, Arnaud de~La~Fortelle, and Masayoshi Tomizuka.
\newblock {INTERACTION} {Dataset}: {An} {INTERnational}, {Adversarial} and {Cooperative} {moTION} {Dataset} in {Interactive} {Driving} {Scenarios} with {Semantic} {Maps}.
\newblock {\em arXiv:1910.03088 [cs, eess]}, September 2019.

\end{thebibliography}

\end{document}